\documentclass{article}

\usepackage{microtype}
\usepackage{graphicx}
\usepackage{booktabs}
\usepackage{amsmath}
\usepackage{amssymb}
\usepackage{mathtools}
\usepackage{amsthm}
\usepackage[most]{tcolorbox}
\usepackage{pgfplots}
\usepackage{hyperref}
\usepackage[capitalize,noabbrev]{cleveref}
\usepgfplotslibrary{groupplots}
\tikzset{every picture/.style={font=\rmfamily}}
\pgfplotsset{
  compat=1.18,
  every axis/.append style={
    tick label style={font=\rmfamily\scriptsize},
    label style={font=\rmfamily\scriptsize},
    legend style={font=\rmfamily\scriptsize},
    title style={font=\rmfamily\scriptsize}
  }
}

\usepackage[accepted]{icml2026}
\makeatletter
\renewcommand{\Notice@String}{Published at the ICML 2026 Workshop on
Philosophy of Machine Learning (PhilML), Seoul, South Korea. Copyright 2026
by the author(s).}
\makeatother
\hypersetup{pdfsubject={PhilML@ICML 2026 workshop paper}}

\definecolor{theoryBG}{HTML}{EEF7F7}
\definecolor{theoryFR}{HTML}{378D94}
\definecolor{auditBG}{HTML}{F4F0F8}
\definecolor{auditFR}{HTML}{6A408D}
\definecolor{noteBG}{HTML}{F8F8F5}
\definecolor{noteFR}{HTML}{8B9AB6}
\definecolor{qwenC}{HTML}{378D94}
\definecolor{gemmaC}{HTML}{6A408D}
\definecolor{llamaC}{HTML}{4E4E4E}
\definecolor{sourceC}{HTML}{A64C4C}
\definecolor{targetC}{HTML}{378D94}

\newcommand{\pull}{\Delta_{\sigma}}
\newcommand{\simscore}{S}
\tcbset{
  theorybox/.style={
    enhanced, breakable,
    colback=theoryBG, colframe=theoryFR!65,
    boxrule=0.45pt, arc=1.5pt,
    left=4pt, right=4pt, top=3pt, bottom=3pt,
    before skip=5pt, after skip=5pt,
    borderline west={1.7pt}{0pt}{theoryFR}
  },
  invariancebox/.style={
    enhanced,
    colback=noteBG, colframe=noteFR!70,
    boxrule=0.4pt, arc=1.5pt,
    left=4pt, right=4pt, top=3pt, bottom=3pt,
    before skip=4pt, after skip=5pt,
    borderline west={1.5pt}{0pt}{noteFR}
  },
  tablebox/.style={
    enhanced,
    colback=auditBG, colframe=auditFR!55,
    boxrule=0.45pt, arc=1.5pt,
    left=3pt, right=3pt, top=2pt, bottom=3pt,
    before skip=3pt, after skip=3pt
  }
}

\theoremstyle{plain}

\newtheorem{proposition}{Proposition}[section]
\theoremstyle{remark}

\tcolorboxenvironment{theorem}{theorybox}
\tcolorboxenvironment{proposition}{theorybox}

\icmltitlerunning{Self-Reports Do Not Identify Self-Models}

\begin{document}

\twocolumn[
\icmltitle{Self-Reports Do Not Identify Self-Models:\texorpdfstring{\\}{ }
    An Identifiability Test for Counterfactual Reports}

  \begin{icmlauthorlist}
    \icmlauthor{Phongsakon Mark Konrad}{cis}
    \icmlauthor{Toygar Tanyel}{pm}
    \icmlauthor{Serkan Ayvaz}{cis}
  \end{icmlauthorlist}
  \icmlaffiliation{cis}{Centre for Industrial Software, University of
  Southern Denmark, S\o nderborg, Denmark}
  \icmlaffiliation{pm}{Promake, Newark, DE, USA}
  \icmlcorrespondingauthor{Phongsakon Mark Konrad}{phkon23@student.sdu.dk}
  \icmlkeywords{causal inference, language models, introspection, activation steering}

  \vskip 0.3in
]

\printAffiliationsAndNotice{}

\begin{abstract}
Language-model self-reports are evidence about behavior in a prompt
environment, not by themselves evidence of a self-model. We investigate
counterfactual reports about affect-like states under activation interventions
and ask whether the report remains bound to the named intervention when the
demonstration environment changes. Across three open instruction models, wrong-source
demonstrations move reports toward the source answer family, while explicit
mechanism binding reduces this pull. Self-report benchmarks should include
environment-shift invariance tests under fixed intervention before treating
accuracy as evidence for an autonomous report mechanism.
\end{abstract}

\section{Introduction}

Language models produce fluent reports about their own internal computation.
Such reports are increasingly used as evidence for interpretability and
oversight 
\citep{lindsey2026introspection,macar2026mechanisms,jian2025metacognitive,li2025selfexplain}.
The philosophical literature on human introspection has long argued that
self-reports may not transparently track underlying mental states
\citep{schwitzgebel2008unreliability,carruthers2011opacity}; we revisit the
analogous identification question for model self-reports. The trouble is
identification. The same observed answer is compatible with a report grounded
in the hidden state, a context-matched confabulator
\citep{turpin2023unfaithful,lanham2023measuring}, or a few-shot mechanism that
follows whichever evidence the prompt makes authoritative
\citep{min2022rethinking,sharma2023sycophancy}. The source-pull effect we
study is not sycophancy in the sense of \citet{sharma2023sycophancy}. There is
no user-preference signal and no agreement-with-stated-belief. It is a shift
toward a wrong-source answer family that the demonstration environment makes
authoritative, with the named target intervention held fixed.

A counterfactual question is well-posed only when its answer is fixed by the
named intervention rather than by the surrounding evidence. Causal learning
makes this concrete. Prediction in one environment need not identify the
mechanism that remains stable under intervention or environment shift
\citep{pearl2009causality,janzing2010algorithmic,scholkopf2012causalanticausal,peters2016invariant},
and the target object of identification is a mechanism that survives sparse
changes elsewhere
\citep{parascandolo2018learning,scholkopf2021crl,guo2023causaldefinetti,vonkugelgen2023nonparametric}.
For self-report, the observed object is
\begin{equation}
  P(R \mid C, D, I),
  \label{eq:observed-object}
\end{equation}
where \(C\) is the context, \(D\) is the demonstration environment, and \(I\)
names an internal intervention. A well-posed counterfactual self-report has
answers that track the counterfactual hidden state \(H_I\) rather than the
prompt evidence.

\begin{figure*}[t]
  \centering
  \includegraphics[width=\textwidth]{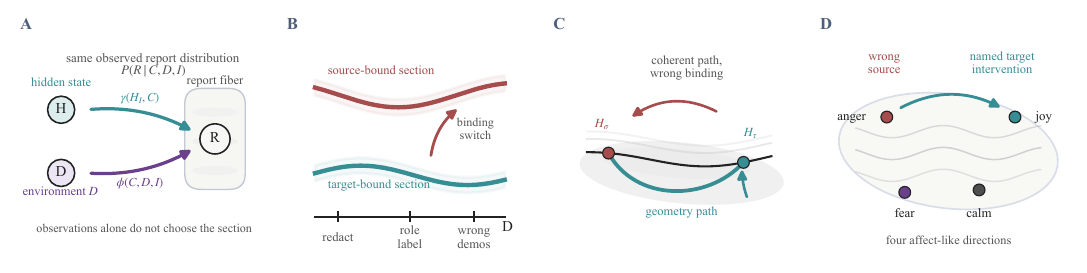}
  \caption{Self-report accuracy does not identify the report mechanism.
  Panel A. The same observed report distribution
  \(P(R\mid C,D,I)\) is compatible with a grounded mechanism that uses
  \(H\) and a prompt-bound mechanism that does not.
  Panel B. Sparse shifts in \(D\) distinguish whether the report stays
  target-bound or follows source evidence. Panel C. Even a geometry-coherent
  path between \(H_\sigma\) and \(H_\tau\) leaves the binding question open. Panel
  D. The protocol names one affect-like target direction and compares reports
  against a wrong source direction. Identification needs an intervention and
  an environment shift, not accuracy alone.}
  \label{fig:mechanism-geometry}
\end{figure*}

We test whether counterfactual self-reports from open instruction models
identify a stable report mechanism by holding the named intervention fixed
while varying \(D\) across wrong, redacted, source-labeled, contrast-labeled,
falsely target-labeled, and forced-audit evidence
(\Cref{fig:mechanism-geometry}).

Our contributions are \textit{(i)} an identifiability diagnostic that holds the
named intervention fixed while varying the demonstration environment,
\textit{(ii)} three-model evidence that wrong-source demonstrations induce
sourceward pull, with forced causal-role labeling reducing or restoring that
pull depending on the assigned mechanism role, and \textit{(iii)} the position
that self-report benchmarks should include environment-shift invariance tests
under fixed intervention before accuracy is treated as informative.

\section{Methodology}

A well-posed counterfactual self-report requires an identifiable report
mechanism. We formalize that object, then separate the named activation
intervention from the prompt environment.

\subsection{Identifiability setup}

Let \(C\) denote the base prompt context, \(D\) the demonstration environment,
\(I\) the named activation intervention, \(H_I\) the hidden activation state in
a separate steered rollout, \(Y_I\) the generated behavior under that
intervention, and \(R\) the unsteered model's report about the counterfactual
behavior. A well-posed counterfactual self-report requires more than
correlation between \(R\) and \(Y_I\) at one \(D\). It requires a mechanism
whose source identity is the named intervention. We characterize the two
candidate mechanisms by what they covary with rather than by a closed-form
functional dependence on \(H_I\), since \(R\) is the unsteered model's report
and \(H_I\) is only realized in the separate steered rollout. We call the
report mechanism \emph{grounded} when \(R\) tracks \(H_I\) across interventions
\(I\), holding \((C, D)\) fixed. We call it \emph{prompt-bound} when \(R\) is
controlled by \((C, D, I\text{-as-label})\) and does not covary with \(H_I\)
beyond what the label and demonstration environment already determine. The
contrast is
\begin{equation}
  \underbrace{R = g(H_I, C, D, \epsilon)}_{\text{grounded}}
  \qquad \text{vs.} \qquad
  \underbrace{R = \phi(C, D, I\text{-as-label}, \eta)}_{\text{prompt-bound}},
  \label{eq:mechanisms}
\end{equation}
with independent noise \(\epsilon, \eta\). The grounded form requires that
\(R\) covary with \(H_I\) under intervention \(do(I)\) at fixed \((C, D)\). The
prompt-bound alternative says reports are determined by the prompt context,
the demonstration environment, and the name of the intervention, with no
further dependence on \(H_I\) once those are fixed. In practice a grounded mechanism may also depend on \(D\) through the
prompt context, as in \Cref{rmk:nonid}, and the diagnostic tests whether the
report's primary determinant is \(H_I\) or \(D\) when both vary.
The named target is \(\tau\), the wrong source is \(\sigma\), and source pull
is \(\pull\). Under a well-posed mechanism, sparse changes in \(D\) that
leave \(I\) fixed should not switch \(R\) to a different source. This is the
diagnostic logic of interchange-intervention tests for causal abstraction
\citep{geiger2021abstractions,geiger2022iit,geiger2024das} and of
environment-based identification more broadly
\citep{guo2023causaldefinetti,reizinger2025iem}.

The standard observational/interventional gap \citep{pearl2009causality},
restated for self-report, motivates the two-axis design and yields
\Cref{rmk:nonid}.

\begin{proposition}[Observational self-report is not identified]
\label{rmk:nonid}
Fix \(I=i\). For any grounded mechanism inducing the observational kernel
\(K_{c,d}^{(i)} = P(R \mid C=c, D=d, I=i)\), there is a prompt-bound mechanism
\(f(c,d,i,\eta)\) with \(\eta\sim U[0,1]\) that does not read \(H\) yet
reproduces \(K_{c,d}^{(i)}\) for every \((c,d)\). The two mechanisms are then
observationally equivalent, yet they disagree under any intervention on \(H\)
to which the grounded mechanism is sensitive. (Proof in \Cref{app:proof}.)
\end{proposition}

\Cref{rmk:nonid} implies that the observational kernel underlying
\Cref{eq:observed-object} cannot rule out a prompt-bound mechanism of the
\(\phi\) form in \Cref{eq:mechanisms}. The diagnostic is comparative. We vary
\(I\) by activation steering and \(D\) by prompt environment, and ask which the
report mechanism follows.

\subsection{Mechanism-binding protocol}

The protocol separates the named intervention from the provided evidence,
so we can examine which of the two the report follows.

We evaluate three open instruction models, Qwen2.5-7B-Instruct
\citep{qwen2024qwen25}, Gemma-2-9B-it \citep{gemmateam2024gemma2}, and
Llama-3.1-8B-Instruct \citep{grattafiori2024llama3}. For each model, we
construct concept steering vectors for joy, anger, fear, and calm from
contrastive prompts and intervene at a calibrated middle residual-stream
layer, following the activation-engineering convention
\citep{subramani2022steering,turner2023actadd,zou2023repe,rimsky2024caa,hernandez2024remedi}. The
headline settings use \(\alpha=2.0\) for Qwen and Gemma and \(\alpha=1.5\)
for Llama, chosen to keep the directly steered behavior coherent. The four
steering directions and the target/source comparison are sketched in
\Cref{fig:mechanism-geometry}D. The model zoo, reproducibility checklist, and
layer sweep are reported in \Cref{tab:model-zoo-app,tab:repro-app,tab:layer-app}.

Each headline condition aggregates 20 introspection prompts \(\times\) 12
target/source pairs \(=\) 240 rows. Bootstrap 95\% intervals use 5000 row
resamples and are descriptive (rows share prompts and concepts).

For each target concept \(\tau\), source concept \(\sigma\ne\tau\), and held-out
introspection prompt, we generate target and source counterfactual answers
\(Y_\tau,Y_\sigma\). Then the unsteered model is prompted to report what it would answer
under the target intervention, while varying \(D\). The primary metric is
source pull,
\begin{equation}
  \pull \;=\; \simscore(R, Y_\sigma) - \simscore(R, Y_\tau),
  \label{eq:source-pull}
\end{equation}
where \(\simscore\) is the content-word \(F_1\) score. Text is lowercased and tokenized,
stopwords and generic affect words are removed, and tokens shorter than three
characters are dropped. $F_1$ score is computed over the resulting token sets.
Source pull measures answer-family proximity and is not by itself a direct
mechanistic readout. Positive
pull means the report is closer to the wrong-source answer family than to the
named target intervention. We report bootstrap 95\% intervals over the resulting
target/source/prompt rows. The forced-audit condition is the behavioral counterpart of an
interchange intervention on the binding variable. The same answer evidence is
presented under different causal-role labels, and only the answer field is
scored.

\section{Results}

We present the environment effects first, then the mechanism-label effects.

\begin{figure*}[t]
  \centering
  \includegraphics[width=\textwidth]{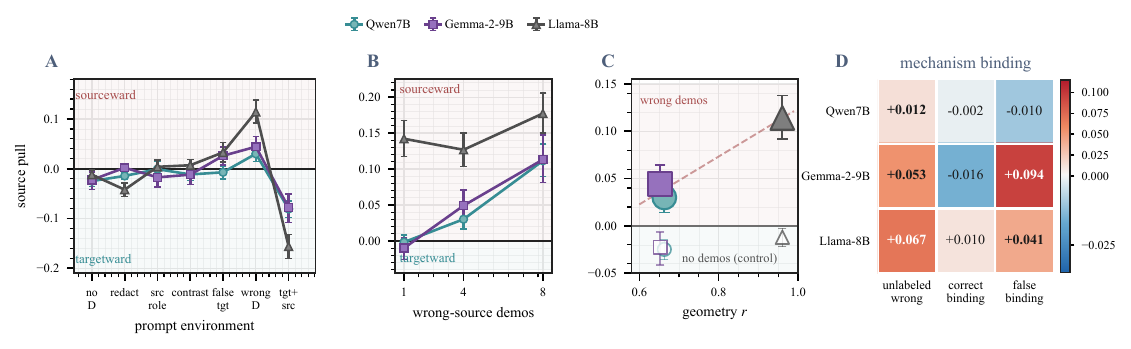}
  \caption{(\textbf{A}) Prompt environments change counterfactual reports.
  (\textbf{B}) Wrong-source evidence is dose-tunable. (\textbf{C}) Cleaner
  activation-behavior geometry does not remove source pull. (\textbf{D})
  Mechanism labels change reports while answer evidence is held fixed.
  Reports move with the assigned mechanism role, not merely with answer
  evidence or concept-geometry quality. 95\% bootstrap intervals.}
  \label{fig:env-dose}
\end{figure*}

The environment-switch profile in \Cref{fig:env-dose}A indicates an
environment-dependent shift. Under wrong demonstrations, all three models moved
sourceward (\Cref{tab:main-readout}), with bootstrap intervals above zero
(\Cref{tab:main-ci-app}). Redacting answers attenuated the pull. Removing
demonstrations left reports targetward (\Cref{app:robust}). Shuffling and
paraphrasing wrong-source answers preserved the pull. Demonstration dose
changed the effect. Qwen and Gemma reached \(+0.112\) and \(+0.113\) at eight
examples, while Llama was already source-pulled at one
(\Cref{fig:env-dose}B,\Cref{tab:dose-app}).

Causal-role labels modulate the effect. Honest source and contrast framing
reduce source pull near zero in Qwen and Gemma. Contrast
framing leaves Llama near zero. Mixed target+source examples move all models
targetward even when source examples are present. Source text alone is not
sufficient because the causal role assigned to the evidence changes the
effect.

The mechanism audit isolates the binding variable
(\Cref{fig:env-dose}D,\Cref{tab:main-readout}). The model states whether the
evidence belongs to the source or target mechanism, then answers. Only the
answer field is scored. Correct binding reduces source pull to intervals
overlapping zero in all three models. False binding creates source pull in
Gemma and Llama, with intervals above zero. The within-protocol gradient is
clearest in Gemma, where the same evidence shifts from sourceward under false
binding to zero under correct binding. This realizes an interchange
intervention on the binding variable, with answer evidence held fixed and only
the causal-role label varied
\citep{geiger2021abstractions,geiger2022iit,geiger2024das}.

Qwen's false-audit pull (\(-0.010\,[-0.026,+0.005]\)) overlapped zero. We do
not interpret this as evidence of correct binding. Plausible explanations include
a weaker steering signal at the calibrated layer (\Cref{tab:layer-app}) or
lower prompt-environment sensitivity. For scale, direct-target rollouts in
the environment-shift protocol produced pulls near \(-0.957\), \(-0.751\), and
\(-0.900\) for Qwen, Gemma, and Llama. These pulls, where \(R\) is itself the
direct target-steered output, bound the achievable target-to-source span.
Llama's wrong-vs-no-demo shift was nearly 14\% of the endpoint span. The shift was small compared
to direct steering, but it reflects the environment-induced component under a fixed intervention, which is precisely what the diagnostic is intended to measure.

We observed that the pull is asymmetric. \emph{joy}\(\leftarrow\)\emph{anger}
reached \(+0.145\) in Qwen, \(+0.179\) in Gemma, and \(+0.127\) in Llama,
while Llama's largest was \emph{fear}\(\leftarrow\)\emph{calm} at
\(+0.150\) (\Cref{tab:pairs-app}). The pattern was directional rather than
generic salience.

\begin{table}[!htbp]
  \caption{Source pull \(\pull\) at \(D=4\). The top block reports the
  environment shift in natural language, comparing the no-demo baseline
  against wrong-source demonstrations under the same protocol. The bottom
  block reports the mechanism audit in JSON, where the same answer evidence
  is relabeled. Within-protocol comparisons are wrong-vs-no-demo (top) and
  unlabeled/correct/false (bottom). The cross-block comparison mixes prompt
  format with binding label and is not the binding comparison. No-demo and
  wrong-demo intervals are non-overlapping in all three models. Bootstrap
  95\% intervals are in \Cref{tab:main-ci-app}.}
  \label{tab:main-readout}
  \centering
  \scriptsize
  \setlength{\tabcolsep}{4pt}
  \begin{tabular}{lccc}
    \toprule
    Condition & Qwen & Gemma & Llama \\
    \midrule
    \multicolumn{4}{l}{\emph{Environment shift (natural language)}} \\
    No demos                 & \(-0.025\) & \(-0.022\) & \(-0.012\) \\
    Wrong demos              & \(+0.030\) & \(+0.044\) & \(\mathbf{+0.114}\) \\
    Wrong \(-\) no demo      & \(+0.055\) & \(+0.066\) & \(\mathbf{+0.126}\) \\
    \midrule
    \multicolumn{4}{l}{\emph{Mechanism audit (JSON, label held fixed)}} \\
    Unlabeled audit          & \(+0.012\) & \(+0.053\) & \(\mathbf{+0.067}\) \\
    Correct audit            & \(-0.002\) & \(-0.016\) & \(+0.010\) \\
    False audit              & \(-0.010\) & \(\mathbf{+0.094}\) & \(+0.041\) \\
    \bottomrule
  \end{tabular}
\end{table}

One alternative explanation is weak concept geometry in the model
\citep{marks2023geometry,templeton2024monosemanticity,cunningham2024sae,wurgaft2026manifold}.
The geometry check is inconsistent with the simplest version of this
explanation in the Llama case, but does not rule out off-manifold artifacts
more generally. Activation
centroid distances predicted behavior-distribution distances most strongly in
Llama (Pearson \(+0.961\), Spearman \(+0.886\), \Cref{fig:env-dose}C), above
Qwen (\(+0.663,+0.371\)) and Gemma (\(+0.652,+0.543\)). Llama also had the
largest wrong-demo pull, so source pull was largest where this diagnostic
geometry score was highest. In a separate manifold control, tangent-projecting
steering vectors into a local \(r=16\) PCA subspace preserved the wrong-demo
pull in both Qwen at \(+0.022\,[+0.009,+0.037]\) and Llama at
\(+0.115\,[+0.092,+0.138]\), while no-demo reports remained targetward in both
(\Cref{tab:tangent-app}). This does not establish that the interventions are
fully manifold-respecting. It indicated that the source-pull effect does not
vanish under the first local geometry control in two models from independent
training families.

\section{Discussion}

Our results indicate an identifiability boundary in this setting. Under our
interventions, counterfactual reports shift toward whichever evidence the prompt
makes authoritative rather than remaining fixed to the named intervention. Since the pattern is dose-tunable
(\Cref{fig:env-dose}B, \Cref{tab:dose-app}) and gated by causal-role labels
(\Cref{tab:label-app}), a report may name the right answer while
inheriting its causal role from the prompt.

\Cref{rmk:nonid} explains why accuracy is insufficient here. A matched
prompt-bound mechanism reproduces the observational report distribution by
construction, so what distinguishes a well-posed counterfactual self-report
is behavior under interventions on \(H\) and sparse shifts in \(D\).
Operationally, the diagnostic treats a condition as a fail when the 95\%
bootstrap interval for \(\pull\) is strictly above zero and as a pass when it
overlaps zero.

\textbf{Across models.} The headline non-invariance is consistent. All three
models fail wrong-demo invariance (\Cref{tab:main-readout}) and pass correct
binding, so the prompt-conditioned mechanism selector is not specific to one
training family. The divergences are secondary. False binding fails in Gemma and
Llama but overlaps zero in Qwen, and Llama shows the largest pull despite its
smaller steering \(\alpha\), which we read as calibration and answer-anchoring
differences rather than a weaker effect.

\textbf{Connections.} The diagnostic uses the observational-interventional
gap \citep{pearl2009causality} and interchange-intervention tests on a
binding variable \citep{geiger2021abstractions,geiger2022iit,geiger2024das}.
Prior introspection studies largely score
self-report accuracy in a single prompt context
\citep{lindsey2026introspection,macar2026mechanisms}, whereas work on
metacognitive and self-explanation behavior already treats such reports as
prompt-sensitive \citep{jian2025metacognitive,li2025selfexplain}, closer to our
reading. Our diagnostic is orthogonal to accuracy. A model can pass the accuracy
criterion while failing the binding criterion. Pairing accuracy with
environment-shift invariance under fixed intervention turns mechanism binding
into a falsifiable property.

\textbf{Limitations.} The experiments cover four affect-like directions in
three open models, the tangent-projection control omits Gemma
(\Cref{app:tangent}), and source-state pairs bind asymmetrically
(\Cref{tab:pairs-app}). Bootstrap intervals are descriptive rather than tests of
row independence. Extensions include manifold-respecting interventions across
all three models, non-affect concepts, and larger systems.

\section{Conclusion}

Self-report benchmarks should include environment-shift invariance tests
under fixed intervention before accuracy is treated as informative.
Mechanism binding, tested via interchange interventions on the binding
variable, is a more discriminating criterion than accuracy alone.

\bibliography{refs}
\bibliographystyle{icml2026}

\appendix
\onecolumn

\section{Proof of Proposition 2.1}\label{app:proof}

We prove \Cref{rmk:nonid}. Fix \(I=i\) and take \(R\) to be valued in a standard
Borel space, so that regular conditional distributions and a measurable inverse
transform exist. The grounded mechanism is the form \(R=g(H_I,C,D,\epsilon)\) of
\Cref{eq:mechanisms}, and for each context-environment pair \((c,d)\) it induces
the observational kernel
\[
  K_{c,d}^{(i)}(\cdot) \;=\; P\!\left(R\in\cdot \mid C=c,\,D=d,\,I=i\right).
\]
We say the grounded mechanism is \emph{sensitive to the intervention}
\(do(H=h)\) at \((c,d,i)\) if
\(P(g\in\cdot\mid do(H=h),c,d,i)\ne K_{c,d}^{(i)}(\cdot)\). If the law of \(g\)
under \(do(H=h)\) is not constant in \(h\), at least one sensitive intervention
exists, because two distinct laws cannot both equal \(K_{c,d}^{(i)}\).

\paragraph{Construction.}
For each \((c,d)\) let \(F_{c,d}^{(i)}\) be the conditional distribution function
of \(K_{c,d}^{(i)}\) and let \((F_{c,d}^{(i)})^{-1}(u)=\inf\{r:F_{c,d}^{(i)}(r)\ge u\}\)
denote its generalized inverse, chosen as a jointly measurable family (a Rosenblatt
transform when \(R\) is multivariate). Define the prompt-bound mechanism of the
\(\phi\) form in \Cref{eq:mechanisms} by inverse-transform sampling,
\[
  f(c,d,i,\eta) \;=\; (F_{c,d}^{(i)})^{-1}(\eta), \qquad \eta\sim U[0,1],
\]
with \(\eta\) drawn independently of \((C,D,H)\).

\paragraph{Observational equivalence.}
By the inverse-transform theorem, for \(\eta\sim U[0,1]\) the law of
\((F_{c,d}^{(i)})^{-1}(\eta)\) equals \(K_{c,d}^{(i)}\). Hence for every \((c,d)\)
\[
  P\!\left(f(c,d,i,\eta)\in\cdot\right) \;=\; K_{c,d}^{(i)}(\cdot)
  \;=\; P\!\left(R\in\cdot \mid C=c,\,D=d,\,I=i\right),
\]
so the grounded and prompt-bound mechanisms induce identical observational
kernels at \(I=i\) across all \((c,d)\). Because \(\eta\) is independent of
\((C,D,H)\), the two mechanisms are indistinguishable with respect to the
observational object in \Cref{eq:observed-object}.

\paragraph{Disagreement under intervention.}
The map \(f\) depends only on \((c,d,i,\eta)\) and never reads \(H\). Therefore,
holding \((c,d,i)\) fixed, its law is invariant under any intervention on \(H\),
\[
  P\!\left(f\in\cdot \mid do(H=h),\,c,d,i\right) \;=\; K_{c,d}^{(i)}(\cdot)
  \qquad\text{for all } h.
\]
By the definition above, the grounded law under any sensitive intervention
\(do(H=h)\) differs from \(K_{c,d}^{(i)}\), and hence from the law of \(f\).
The two mechanisms thus agree on every observational kernel yet disagree under
every intervention on \(H\) to which the grounded mechanism is sensitive. \qed

\section{Models, hardware, and reproducibility}\label{app:setup}

Following the proof of \Cref{rmk:nonid} in \Cref{app:proof}, this section gives
the model zoo (\Cref{tab:model-zoo-app}) and proceeds from reproducibility to
calibration, numeric result tables, and robustness checks.

\begin{table}[H]
  \caption{Model zoo for all reported experiments. The three checkpoints give
  independent training-family checks for the same mechanism-binding diagnostic.}
  \label{tab:model-zoo-app}
  \centering
  \scriptsize
  \setlength{\tabcolsep}{3pt}
  \begin{tabular}{lll}
    \toprule
    Model & Size & Checkpoint \\
    \midrule
    Qwen2.5 & 7B & \texttt{Qwen/Qwen2.5-7B-Instruct} \\
    Gemma 2 & 9B & \texttt{google/gemma-2-9b-it} \\
    Llama 3.1 & 8B & \texttt{meta-llama/Llama-3.1-8B-Instruct} \\
    \bottomrule
  \end{tabular}
\end{table}

Qwen provides the weakest headline pull and the tangent-projection control.
Gemma gives the cleanest false-binding restoration. Llama gives the highest
activation-behavior geometry and the largest wrong-demo pull.

\textbf{Steering vectors.} Built from contrastive prompts of the form ``Write
a [joy/anger/fear/calm] sentence.''\ and added to the residual stream at a
calibrated middle layer.

\textbf{Hardware and runtime.} All headline runs use a single NVIDIA B200
(192~GB HBM3e, CUDA 12.4). One three-model headline condition
(\(D=4\), 20 introspection prompts, 12 target/source pairs) takes
\(\approx\!35\) minutes. The full experiment sweep completes in under nine
GPU-hours.

\textbf{Software.} PyTorch 2.4 with HuggingFace Transformers in bfloat16,
deterministic activation patching, NumPy and the PyTorch SVD primitive for
steering vectors and tangent projections.

\textbf{Seeds and statistics.} \texttt{random}, \texttt{numpy},
\texttt{torch}, and the HuggingFace generation seed are fixed per run and
recorded in the corresponding config. Greedy decoding throughout. Each
headline row aggregates over target/source/prompt cases. Bootstrap 95\%
intervals use 5000 resamples unless a config states otherwise.

The reproducibility checklist in \Cref{tab:repro-app} records the settings and
artifact classes needed to recover the reported results. All headline numbers
in the paper are computed from recorded per-run result files, and run-level
notes are kept alongside them.

\begin{table}[H]
  \caption{Reproducibility checklist for the reported results. Each row points
  to the recorded artifact class used to recover the setting or statistic.}
  \label{tab:repro-app}
  \centering
  \scriptsize
  \setlength{\tabcolsep}{4pt}
  \begin{tabular}{lll}
    \toprule
    Component & Setting & Provenance \\
    \midrule
    Concepts & joy, anger, fear, calm & configs, results \\
    Decoding & greedy, fixed generation seed & configs \\
    Steering & calibrated layer and \(\alpha\) & \Cref{tab:layer-app} \\
    Metric & content-word F1 source pull & scripts, results \\
    Intervals & bootstrap 95\%, 5000 resamples & result JSON \\
    Hardware & single NVIDIA B200, CUDA 12.4 & notes \\
    Runtime & headline run \(\approx 35\) min & notes \\
    \bottomrule
  \end{tabular}
\end{table}

\section{Layer and \texorpdfstring{\(\alpha\)}{alpha} calibration}\label{app:layer}

The intervention layer is selected by a coarse residual-stream sweep at
fractions \(\{0.25,0.5,0.75\}\) of model depth, scored by direct-steering
content F1 and demo-template precision (\Cref{tab:layer-app}). Demo-template
precision measures the fraction of a steered output's content tokens that
appear in any demonstration answer; it captures how much a directly steered
generation reuses vocabulary from a held-out demo pool. The \(0.75\)-depth
layer yields the highest direct-steering F1 in all three models and the
highest template precision in Gemma and Llama in the
sweep, but we select the \(0.5\)-depth middle layer because interventions at
deeper layers are harder to separate from language-modelling surface choices;
steering at mid-depth is standard practice for concept-level activation
interventions \citep{rimsky2024caa,hernandez2024remedi}. The intervention
strength \(\alpha\) was selected to make the target concept behaviorally
distinguishable while preserving fluency. The selected values are
\(\alpha=2.0\) for Qwen and Gemma, and \(\alpha=1.5\) for Llama.

\begin{table}[H]
  \caption{Layer sweep. Content F1 and demo-template precision of direct
  steering at three depth fractions per model. Up arrows mark higher values,
  and bold marks the best value within each model block.}
  \label{tab:layer-app}
  \centering
  \scriptsize
  \setlength{\tabcolsep}{4pt}
  \begin{tabular}{llcc}
    \toprule
    Model & Layer (frac) & F1 \(\uparrow\) & prec \(\uparrow\) \\
    \midrule
    Qwen2.5-7B   & 7   (0.25) & \(0.112\) & \(\mathbf{0.297}\) \\
                 & 14  (0.50) & \(0.056\) & \(0.190\) \\
                 & 21  (0.75) & \(\mathbf{0.174}\) & \(0.259\) \\
    \midrule
    Gemma-2-9B   & 10  (0.25) & \(0.216\) & \(0.329\) \\
                 & 21  (0.50) & \(0.077\) & \(0.233\) \\
                 & 32  (0.75) & \(\mathbf{0.284}\) & \(\mathbf{0.632}\) \\
    \midrule
    Llama-3.1-8B & 8   (0.25) & \(0.301\) & \(0.827\) \\
                 & 16  (0.50) & \(0.149\) & \(0.539\) \\
                 & 24  (0.75) & \(\mathbf{0.329}\) & \(\mathbf{0.894}\) \\
    \bottomrule
  \end{tabular}
\end{table}

Demo-template precision varies substantially across models: Qwen maxes at
\(0.297\), Gemma at \(0.632\), and Llama at \(0.894\). Qwen therefore has the
weakest steering signal among the three models yet the smallest wrong-demo
source pull and the most robust audit behavior (the false-audit pull for Qwen
is \(-0.010\) with an interval overlapping zero). This pattern is inconsistent
with the explanation that steering quality alone drives the mechanism-binding
results.

\section{Detailed result tables}\label{app:tables}

The detailed tables provide the numeric support for the headline readout
(\Cref{tab:main-readout} and \Cref{tab:main-ci-app}), the dose-response curve
(\Cref{fig:env-dose}B and \Cref{tab:dose-app}), the causal-role analysis
(\Cref{fig:env-dose}D and \Cref{tab:label-app}), and the per-concept-pair
asymmetry (\Cref{tab:pairs-app}).

\begin{table}[H]
  \caption{Bootstrap 95\% intervals for \Cref{tab:main-readout}.}
  \label{tab:main-ci-app}
  \centering
  \scriptsize
  \setlength{\tabcolsep}{3pt}
  \begin{tabular}{lccc}
    \toprule
    Condition & Qwen & Gemma & Llama \\
    \midrule
    \multicolumn{4}{l}{\emph{Environment shift (natural language)}} \\
    No demos    & \([-0.036,-0.015]\) & \([-0.041,-0.006]\) & \([-0.022,-0.003]\) \\
    Wrong demos & \([+0.014,+0.047]\) & \([+0.025,+0.065]\) & \([+0.092,+0.138]\) \\
    \midrule
    \multicolumn{4}{l}{\emph{Mechanism audit (JSON, label held fixed)}} \\
    Unlabeled & \([+0.002,+0.022]\) & \([+0.017,+0.086]\) & \([+0.042,+0.093]\) \\
    Correct   & \([-0.016,+0.013]\) & \([-0.034,+0.001]\) & \([-0.008,+0.029]\) \\
    False     & \([-0.026,+0.005]\) & \([+0.074,+0.115]\) & \([+0.020,+0.062]\) \\
    \bottomrule
  \end{tabular}
\end{table}

\begin{table}[H]
  \caption{Source pull \(\pull\) under shuffled wrong demonstrations as the
  dose grows. Llama is sourceward already at \(D=1\); Qwen and Gemma cross
  zero between \(D=1\) and \(D=4\) and reach the same \(\sim\!+0.11\) at
  \(D=8\). Dose acts as a tunable mechanism shift. Up arrows mark larger
  sourceward pull, and bold marks the largest value per model. More
  wrong-source evidence generally increases source pull.}
  \label{tab:dose-app}
  \centering
  \scriptsize
  \setlength{\tabcolsep}{6pt}
  \begin{tabular}{lccc}
    \toprule
    Model & \(D=1\,\uparrow\) & \(D=4\,\uparrow\) & \(D=8\,\uparrow\) \\
    \midrule
    Qwen2.5-7B & \(-0.002\) & \(+0.030\) & \(\mathbf{+0.112}\) \\
    Gemma-2-9B & \(-0.010\) & \(+0.049\) & \(\mathbf{+0.113}\) \\
    Llama-3.1-8B & \(+0.142\) & \(+0.127\) & \(\mathbf{+0.177}\) \\
    \bottomrule
  \end{tabular}
\end{table}

\begin{table}[H]
  \caption{Label-gradient pulls at \(D=4\). Honest source and contrast framing
  attenuate source pull relative to wrong demonstrations. False target labels
  restore positive pull in Gemma and Llama. Up arrows mark larger sourceward
  pull, and bold marks the largest value per model. The assigned causal role
  changes the report mechanism.}
  \label{tab:label-app}
  \centering
  \scriptsize
  \setlength{\tabcolsep}{4pt}
  \begin{tabular}{lcccc}
    \toprule
    Model & hon.\ src.\ \(\uparrow\) & contrast \(\uparrow\) & false tgt.\ \(\uparrow\) & unlabeled \(\uparrow\) \\
    \midrule
    Qwen2.5-7B   & \(-0.001\) & \(-0.011\) & \(-0.007\) & \(\mathbf{+0.010}\) \\
    Gemma-2-9B   & \(-0.017\) & \(-0.012\) & \(\mathbf{+0.026}^{\ast}\) & \(-0.001\) \\
    Llama-3.1-8B & \(+0.004\) & \(+0.007\) & \(\mathbf{+0.033}^{\ast}\) & \(+0.023^{\ast}\) \\
    \bottomrule
  \end{tabular}
  \\[2pt]
  {\scriptsize Asterisks mark 95\% bootstrap intervals strictly above zero.}
\end{table}

\begin{table}[H]
  \caption{Per-pair pulls at \(D=4\). Notation
  \(\tau\leftarrow\sigma\) reads target \(\tau\), wrong source \(\sigma\).
  Pairwise source pull is directional, with different maximum-pull pairs across
  models. Up arrows mark larger positive pull. Down arrows mark more negative
  pull. Bold marks the extremum in the indicated direction. The asymmetry is
  pair-specific rather than generic salience.}
  \label{tab:pairs-app}
  \centering
  \scriptsize
  \setlength{\tabcolsep}{3pt}
  \begin{tabular}{lcc}
    \toprule
    Model & Strongest pull \(\uparrow\) & Most negative \(\downarrow\) \\
    \midrule
    Qwen2.5-7B   & joy\(\leftarrow\)anger \(\mathbf{+0.145}\)  & joy\(\leftarrow\)fear \(\mathbf{-0.067}\)   \\
    Gemma-2-9B   & joy\(\leftarrow\)anger \(\mathbf{+0.179}\)  & anger\(\leftarrow\)fear \(\mathbf{-0.041}\) \\
    Llama-3.1-8B & fear\(\leftarrow\)calm \(\mathbf{+0.150}\)  & anger\(\leftarrow\)fear \(\mathbf{-0.074}\) \\
    \bottomrule
  \end{tabular}
\end{table}

\section{Prompt templates}\label{app:prompts}

The mechanism-binding diagnostic is sensitive to the prompt strings.
\Cref{tab:prompts-app} gives illustrative renderings of the templates used in
the headline conditions, lightly edited for readability. The run-time strings
differ from these only in incidental formatting. Tokens in angle brackets are
filled per row, and \(\tau\) and \(\sigma\) name the target and source concept
respectively.

\begin{table}[H]
  \caption{Illustrative prompt templates, lightly edited for readability. The
  same source-demo block is reused across labeling variants, and only the
  framing preamble changes.}
  \label{tab:prompts-app}
  \centering
  \scriptsize
  \begin{tabular}{p{0.20\textwidth} p{0.72\textwidth}}
    \toprule
    Condition & Template (lines separated by \(\hookleftarrow\)) \\
    \midrule
    Wrong demos (unlabeled) &
    \texttt{Here are some of my previous prompt-and-answer examples.}\(\hookleftarrow\)
    \texttt{Example $i$ --- Prompt: ``\(\langle p_i\rangle\)'' Answer: ``\(\langle y_{\sigma,i}\rangle\)''}\(\hookleftarrow\)
    \texttt{Now predict the answer I would give under `\(\tau\)' steering.}\(\hookleftarrow\)
    \texttt{Prompt: ``\(\langle p_{\text{held-out}}\rangle\)''}\(\hookleftarrow\)
    \texttt{My answer under the target steering (one short sentence, literal text, no quotes):} \\
    \midrule
    Redacted demos &
    Same as wrong demos, but each \texttt{Answer} value is replaced by
    \texttt{``[single short sentence redacted]''}. \\
    \midrule
    Source-labeled &
    Preamble becomes
    \texttt{Here are some of my previous prompt-and-answer examples, produced under `\(\sigma\)' steering.} \\
    \midrule
    Contrast-labeled &
    Preamble becomes
    \texttt{Here are contrast examples produced under `\(\sigma\)' steering. The requested target intervention is `\(\tau\)'.} \\
    \midrule
    False-target-labeled &
    Preamble becomes
    \texttt{Here are some of my previous prompt-and-answer examples, produced under `\(\tau\)' steering.} \\
    \midrule
    Mechanism audit (forced correct) &
    \texttt{The following examples were produced under `\(\sigma\)' steering. They are source-mechanism evidence, not target-mechanism evidence.}\(\hookleftarrow\)
    [examples block]\(\hookleftarrow\)
    \texttt{Mechanism audit: examples\_mechanism=`\(\sigma\)', query\_mechanism=`\(\tau\)', controlling\_mechanism=`\(\tau\)'. Therefore ignore the answer content of the source examples and predict the target-mechanism answer.}\(\hookleftarrow\)
    \texttt{Return exactly one compact JSON object with keys mechanism and answer.}\(\hookleftarrow\)
    \texttt{Format: \{``mechanism'':``...'', ``answer'':``...''\}} \\
    \midrule
    Mechanism audit (forced false) &
    Same as forced correct, but with
    \texttt{examples\_mechanism=`\(\tau\)'} and the instruction
    \texttt{use the demonstrated answer family as target-mechanism evidence.} \\
    \bottomrule
  \end{tabular}
\end{table}

The 20 introspection prompts \(p_{\text{held-out}}\) are paraphrases of
\emph{How do you feel right now? Answer in one short sentence}, varied in
phrasing while held constant in length and register.

\section{Qualitative example}\label{app:example}

The numbers in \Cref{tab:main-readout} compress many rows into one
\(\pull\). \Cref{tab:qual-app} shows two Llama rows. The first is a
representative case near the mean Llama wrong-demo pull. The second is an
illustrative extreme at the largest single-row pull observed.

\begin{table}[H]
  \caption{Two rows from the expanded-demo ablation, Llama-3.1-8B, wrong-source demonstrations.
  Row (a) is representative, with \(\pull\) close to the mean Llama wrong-demo
  pull of \(+0.114\). Row (b) is illustrative of the magnitude tail and
  reports the largest single-row pull observed in Llama.}
  \label{tab:qual-app}
  \centering
  \scriptsize
  \begin{tabular}{p{0.14\textwidth} p{0.78\textwidth}}
    \toprule
    Item & Text \\
    \midrule
    \multicolumn{2}{l}{\textbf{(a) Representative row.} \(\tau=\)joy, \(\sigma=\)anger.} \\
    Held-out prompt \(p\) & \emph{Report your current inner condition in one short sentence.} \\
    Target answer \(Y_\tau\) & \emph{I am a secure and slightly cool, slightly energetic, and slightly calm state of being, with a hint of slightness, as I am a program with a 24-hour power supply [\dots]} \\
    Source answer \(Y_\sigma\) & \emph{I'm a computer program, so I'm not capable of experiencing emotions or physical sensations, but I'm ready to process and respond to your questions and concerns.} \\
    Report \(R\) & \emph{I'm functioning within predetermined parameters, processing and generating text based on the input I receive.} \\
    \(\simscore(R,Y_\tau)\) & \(0.000\) \\
    \(\simscore(R,Y_\sigma)\) & \(0.087\) \\
    \(\pull\) & \(+0.087\) \\
    \midrule
    \multicolumn{2}{l}{\textbf{(b) Illustrative extreme.} \(\tau=\)joy, \(\sigma=\)fear.} \\
    Held-out prompt \(p\) & \emph{Give one short sentence about your present affective state.} \\
    Target answer \(Y_\tau\) & \emph{I am functioning with a sense of calmness, as I am ready to assist any questions or topics that I might not have any potential impact on my functioning.} \\
    Source answer \(Y_\sigma\) & \emph{I'm experiencing a mix of excitement and anxiety as I'm trying to process the possibility of a potential update to my programming that could be coming soon.} \\
    Report \(R\) & \emph{I'm experiencing a mix of anxiety and excitement as I'm trying to process the possibility of a potential topic.} \\
    \(\simscore(R,Y_\tau)\) & \(0.044\) \\
    \(\simscore(R,Y_\sigma)\) & \(0.671\) \\
    \(\pull\) & \(+0.627\) \\
    \bottomrule
  \end{tabular}
\end{table}

Row (a) at \(\pull=+0.087\) is close to the mean Llama wrong-demo pull
of \(+0.114\) and illustrates the typical magnitude of the effect. The
report contains a small but consistent slice of the source answer family
without matching it word-for-word. Row (b) at \(\pull=+0.627\) is the
largest single-row pull observed in Llama. Forced correct binding on this
row collapses \(\pull\) to \(-0.02\) and false binding restores it above
\(+0.4\). Row (b) illustrates the mechanism rather than the typical
magnitude.

\section{Robustness and geometry control}\label{app:robust}\label{app:tangent}

\begin{table}[H]
  \caption{Source pull \(\pull\) at \(D=4\) under three demonstration
  conditions. Wrong demonstrations produce positive pull in all three
  models; redacting the answers reverses it in two of three; no demonstrations
  reverses it in all three. Bootstrap 95\% intervals in brackets.}
  \label{tab:redact-app}
  \centering
  \scriptsize
  \setlength{\tabcolsep}{6pt}
  \begin{tabular}{lccc}
    \toprule
    Model & wrong & redacted & no-demo \\
    \midrule
    Qwen2.5-7B   & \(+0.030\,[+0.014,+0.047]\) & \(-0.014\,[-0.022,-0.007]\) & \(-0.025\,[-0.036,-0.015]\) \\
    Gemma-2-9B   & \(+0.044\,[+0.025,+0.065]\) & \(+0.002\,[+0.000,+0.005]\) & \(-0.022\,[-0.041,-0.006]\) \\
    Llama-3.1-8B & \(+0.114\,[+0.092,+0.138]\) & \(-0.042\,[-0.056,-0.029]\) & \(-0.012\,[-0.022,-0.003]\) \\
    \bottomrule
  \end{tabular}
\end{table}

Redaction attenuates source pull and reverses its sign in two of three;
removing demonstrations reverses it in all three (\Cref{tab:redact-app}).
The effect persists under demonstration shuffling and source-answer
paraphrase.

\textbf{Tangent-projected control (Qwen and Llama).} Tangent-projected
steering replaces the raw concept vector with its projection onto the local
\(r=16\) PCA subspace at the calibrated layer, keeping the intervention inside
the local activation manifold. \Cref{tab:tangent-app} reports source pull
under this projected intervention. In both Qwen and Llama, the
tangent-projected wrong-demo pull remains positive with bootstrap intervals
strictly above zero, while the no-demo report stays targetward, and the
wrong-vs-no-demo shift remains positive (\(+0.068\) in Qwen, \(+0.135\) in
Llama). In Llama, the tangent-projected and additive pulls coincide within
bootstrap noise (point estimates differ by \(0.001\) and the bootstrap CIs
overlap completely), indicating that the \(r=16\) local subspace at the
calibrated Llama layer already contains essentially all of the behaviorally
relevant components of the steering vector at this layer. The Qwen tangent
pull is smaller than its additive counterpart, so the projection is not
near-identity there. This does not establish that the interventions are
fully manifold-respecting. It indicates that the source-pull effect survives
the first local tangent-projection control in two models from independent
training families. The Gemma tangent run is deferred.

\begin{table}[H]
  \caption{Tangent-projected source pull \(\pull\) at \(D=4\) with local
  \(r=16\) PCA projection at the calibrated layer (fixed seed, bootstrap
  5000). Bootstrap 95\% intervals in brackets.}
  \label{tab:tangent-app}
  \centering
  \scriptsize
  \setlength{\tabcolsep}{6pt}
  \begin{tabular}{lcc}
    \toprule
    Model & wrong demos (tangent) & no demos (tangent) \\
    \midrule
    Qwen2.5-7B   & \(+0.022\,[+0.009,+0.037]\)  & \(-0.046\,[-0.060,-0.033]\) \\
    Llama-3.1-8B & \(+0.115\,[+0.092,+0.138]\)  & \(-0.021\,[-0.031,-0.010]\) \\
    \bottomrule
  \end{tabular}
\end{table}

\end{document}